\documentclass{article}

\usepackage{microtype}
\usepackage{graphicx}
\usepackage{subfigure}
\usepackage{booktabs} 

\usepackage{hyperref}

\usepackage{mydefs}

\usepackage[accepted]{icml2019}

\icmltitlerunning{Stochastic Blockmodels meet Graph Neural Networks}

\begin{document}

\twocolumn[
\icmltitle{Stochastic Blockmodels meet Graph Neural Networks}

\icmlsetsymbol{equal}{*}

\begin{icmlauthorlist}
\icmlauthor{Nikhil Mehta$^*$}{duke}
\icmlauthor{Lawrence Carin}{duke}
\icmlauthor{Piyush Rai}{iitk}
\end{icmlauthorlist}

\icmlaffiliation{duke}{Department of Electrical and Computer Engineering, Duke University}
\icmlaffiliation{iitk}{Department of Computer Science, IIT Kanpur}
\icmlcorrespondingauthor{Nikhil Mehta}{nikhilmehta.dce@gmail.com}

\icmlkeywords{Machine Learning, ICML}

\vskip 0.3in
]

\printAffiliationsAndNotice{$^*$The major part of this work was done when Nikhil Mehta was at IIT Kanpur.}  

\begin{abstract}
Stochastic blockmodels (SBM) and their variants, $e.g.$, mixed-membership and overlapping stochastic blockmodels, are latent variable based generative models for graphs. They have proven to be successful for various tasks, such as discovering the community structure and link prediction on graph-structured data. Recently, graph neural networks, $e.g.$, graph convolutional networks, have also emerged as a promising approach to learn powerful representations (embeddings) for the nodes in the graph, by exploiting graph properties such as locality and invariance. In this work, we unify these two directions by developing a \emph{sparse} variational autoencoder for graphs, that retains the interpretability of SBMs, while also enjoying the excellent predictive performance of graph neural nets. Moreover, our framework is accompanied by a fast recognition model that enables fast inference of the node embeddings (which are of independent interest for inference in SBM and its variants). Although we develop this framework for a particular type of SBM, namely the \emph{overlapping} stochastic blockmodel, the proposed framework can be adapted readily for other types of SBMs. Experimental results on several benchmarks demonstrate encouraging results on link prediction while learning an interpretable latent structure that can be used for community discovery.
\end{abstract}

\section{Introduction}

\noindent Learning the latent structure in graph-structured data~\cite{fortunato2010community,goldenberg2010survey,schmidt2013nonparametric} is an important problem in a wide range of domains, such as social and biological network analysis and recommender systems. These latent structures help discover the underlying communities in the network, as well as in predicting potential links between nodes. Latent space models~\cite{hoff2002latent} and their structured extensions, such as the stochastic blockmodel~\cite{nowicki2001estimation} and variants like the infinite relational model (IRM)~\cite{kemp2006learning}, mixed-membership stochastic blockmodel (MMSB)~\cite{airoldi2008mixed}, and the overlapping stochastic blockmodel (OSBM)~\cite{miller2009nonparametric,latouche2011overlapping} accomplish this by learning low-dimensional, interpretable node embeddings defined via structured latent variables. These embeddings can be used to identify the community membership(s) of each node in the graph, as well as for tasks such as link prediction.

The overlapping stochastic blockmodel (OSBM), also known as the latent feature relational model (LFRM), is a particularly appealing model for relational data~\cite{miller2009nonparametric,latouche2011overlapping,zhu2016max}. The OSBM/LFRM models each node in the graph as belonging to one or more communities using a binary membership vector, and defines the link probability between any pair of nodes as a \emph{bilinear} function of their community membership vectors. Despite its appealing properties, the OSBM/LFRM has a number of limitations. In particular, although usually considered to be more expressive~\cite{miller2009nonparametric} than models such as IRM and MMSB, a single layer of binary node embeddings and the bilinear model for the link generation can still limit the expressiveness of OSBM/LFRM. Moreover, it has a challenging inference procedure, which primarily relies on MCMC~\cite{miller2009nonparametric,latouche2011overlapping} or mean-field variational inference~\cite{zhu2016max}. Although recent models have tried to improve the expressiveness of OSBM/LFRM, $e.g.$, by assuming a \emph{deep} hierarchy of binary-vector-based node embeddings~\cite{hu2017deep}, inference in such models remains intractable, requiring expensive MCMC-based inference. It is therefore desirable to have a model that retains the basic spirit to OSBM/LFRM ($e.g.$, easy interpretability and strong link prediction performance), but with greater expressiveness, and a simpler and scalable inference procedure.

Motivated by these desiderata, we develop a deep generative framework for graph-structured data, that inherits the easy interpretability of overlapping stochastic blockmodels, but is much more expressive and enjoys a fast inference procedure. Our framework is based on a novel, \emph{sparse} variant of the variational autoencoder (VAE)~\cite{kingma2013auto}, designed to model graph-structured data. Our VAE-based setup comprises a nonlinear generator/decoder for the graph and a nonlinear encoder based on the graph convolutional network (GCN)~\cite{kipf2016semi} (although other graph neural networks can also be used). Our framework posits each node of the graph to have an embedding in the form of a sparse latent representation (modeled by a Beta-Bernoulli process~\cite{griffiths2011indian}, which also enables \emph{learning} the size of the embeddings). The generator/decoder part of the VAE models the probability of a link between two nodes via a nonlinear function (defined by a deep neural network) of their associated embeddings. The encoder part of the VAE consists of a fast \emph{recognition} model that is designed leveraging reparameterization method for Beta and Bernoulli distributions~\cite{MaddisonMT16,ErickSBVAE}. The recognition model, based on stochastic gradient variational Bayes (SGVB) inference, enables fast inference of the node embeddings. In contrast, the traditional stochastic blockmodels rely on iterative MCMC or variational inference procedures for inferring the node embeddings. Consequently, the SGVB inference algorithm we develop is also of independent interest, since the recognition model enables fast inference of the node embeddings in \emph{single-layer} overlapping stochastic blockmodels.

\section{Preliminaries}
We first introduce notation and then briefly describe the overlapping stochastic blockmodel (OSBM)~\cite{latouche2011overlapping,miller2009nonparametric,zhu2016max}. As described in the next section, our deep generative model enriches OSBM by endowing it with a deep architecture based on a \emph{sparse} variational autoencoder, and a fast inference algorithm based on a recognition model. We assume that the graph is given as an adjacency matrix $\Amat \in \{0,1\}^{N \times N}$, where $N$ denotes the number of nodes. We assume $A_{nm} = 1$ if there exist a link between node $n$ and node $m$, and otherwise $A_{nm} = 0$. In addition to $\Amat$, for each node we may also be provided node features. These are given in the form of an $N \times D$ matrix $\Xmat$, with $\xv_n \in {\R}^{D}$ being the node features for node $n$, and $D$ being the number of observed features.


The OSBM~\cite{latouche2011overlapping,miller2009nonparametric,zhu2016max} is a stochastic blockmodel for networks; it assumes each node $n$ has an associated binary vector (node embedding), also termed a \emph{latent feature vector} $\zv_n \in \{0,1\}^K$. Within the node embedding, $z_{nk}=1$ denotes that node $n$ belongs to cluster/community $k$, and $z_{nk} = 0$ otherwise. The OSBM allows each node to simultaneously belong to multiple communities, and defines the link probability between two nodes via a bilinear function of their latent feature vectors 
\beq
p(A_{nm} = 1|\zv_n,\zv_m,\Wmat) = \sigma(\zv_n^\top \Wmat \zv_m)
\eeq
where entry $w_{k\ell}$ in $\Wmat \in \mathbb{R}^{K\times K}$ affects the probability of a link between node $n$ and node $m$ belonging to cluster $k$ and cluster $\ell$, respectively. 

The nonparametric latent feature relational model (LFRM) is a specific type of OSBM, that leverages the Indian Buffet Process (IBP) prior~\cite{miller2009nonparametric} on the $N\times K$ binary matrix $\Zmat = [\zv_1,\ldots,\zv_N]^\top$ of the node-community membership vectors. Use of the IBP enables {\em learning} the number of communities. Inference in LFRM/OSBM is typically performed via MCMC or variational inference~\cite{miller2009nonparametric,latouche2011overlapping,zhu2016max}, which tends to be slow and often cannot scale easily to more than a few hundred nodes.

\section{Deep Generative OSBM}

\begin{figure*}[!htbp]
 \centering     
    \includegraphics[width=0.48\textwidth]{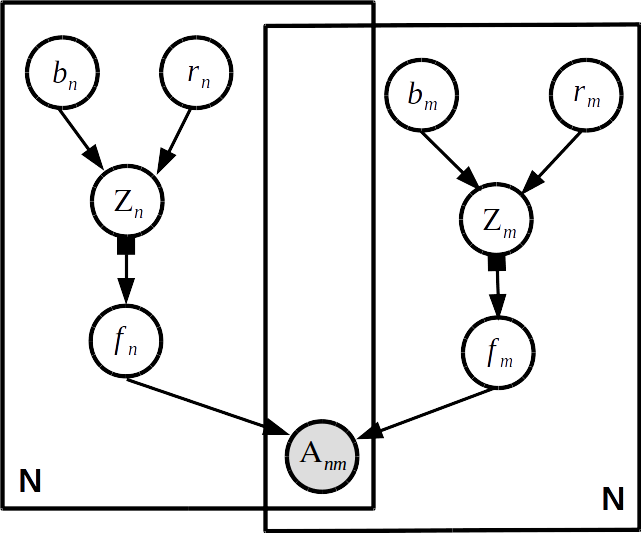}
    \hspace{2em}
    \includegraphics[width=0.45\textwidth]{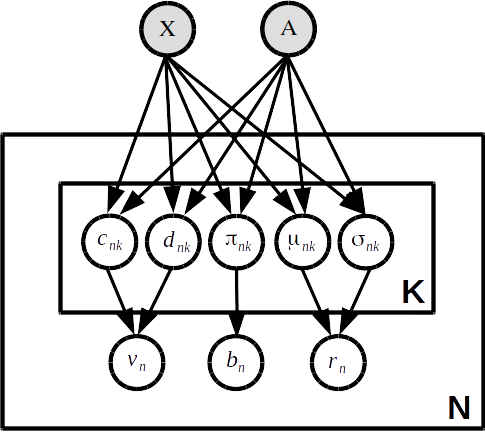}
\caption{\small{(Left) The generator/decoder model in plate notation. Note that the mapping from $\zv_n$ to $\fv_n$ is a deterministic nonlinear transformation, modeled by a deep neural network. (Right) The encoder model, defined by a graph convolutional network~\cite{kipf2016semi} that takes as input the network $\Amat$ and node features $\Xmat$ (if available) and outputs the parameters of the variational distributions of the model parameters.}}
\label{fig:dec-enc}
\end{figure*}

We now present our sparse VAE based deep generative framework for overlapping stochastic blockmodel. The proposed architecture, depicted in Fig.~\ref{fig:dec-enc} (left), associates each link $A_{nm} \in \{0,1\}$ with two latent embeddings $\zv_n$ and $\zv_m$ (for the nodes associated with this link). Each link probability is modeled as a nonlinear function of the embeddings of its associated nodes. Unlike the standard VAE that assumes dense, Gaussian-distributed embeddings, since we wish to use the node embeddings to also infer the community membership(s) of each node (as it is one of the goals of stochastic blockmodels), we impose sparsity on the node embeddings. This is done by modeling them as a sparse vector of the form $\zv_n = \bv_n \odot \rv_n$, where $\bv_n \in \{0,1\}^K$ is a binary vector modeled using a stick-breaking process prior~\cite{pmlr-v2-teh07a} and $\rv_n \in \mathbb{R}^K$ is a real-valued vector with a Gaussian prior. Modeling $\bv_n$ using the stick-breaking prior enables learning the node embedding size $K$ from data. Note that, unlike the OSBM/LFRM, which assumes the node embedding $\zv_n$ to be a strictly binary vector, our framework models it as a sparse real-valued vector, providing a more flexible and informative representation for the nodes. In particular, this enables inference of not just the node's membership into communities, but also the \emph{strength} of the membership in each of the communities the node belongs to. Specifically, $b_{nk} \in \{0,1\}$ denotes whether node $n$ belongs to cluster $k$ or not, and $r_{nk} \in \mathbb{R}$ denotes the strength.

\subsection{VAE Generator/Decoder}

Given the node embeddings $\zv_n = \bv_n \odot \rv_n$, the VAE decoder generates each link in the graph as $A_{nm} \sim p_\theta(\zv_n,\zv_m)$, where probability distribution $p_\theta$ defines a \emph{decoder} or generator model for the graph. This decoder can consist of one or more layers of deterministic \emph{nonlinear} transformation of the node embeddings $\zv_n$. Denoting the overall transformation for a node embedding $\zv_n$ as $f(\zv_n) = \fv_n$,  we model the probability of a link as $p(A_{nm} = 1|\fv_n,\fv_m) = \sigma(\fv_n^\top \fv_m)$, where the nonlinear function $f$ can be modeled by a deep neural network (in our experiments, we use a deep neural net with each hidden layer having leaky ReLU nonlinearity). Figure~\ref{fig:dec-enc} (left) depicts the generator.

We model the binary vector $\bv_n \in \{0,1\}^K$, denoting node-community memberships, using the stick-breaking construction of the IBP~\cite{pmlr-v2-teh07a}, which enables learning of the \emph{effective} $K$ by specifying a sufficiently large truncation level $K$. The stick-breaking construction is given as follows
\beqs
  v_k &\sim& \text{Beta}(\alpha,1), \quad k=1,\ldots,K  
  \\ \pi_k &=& \prod_{j=1}^k v_j,\quad b_{nk} \sim \text{Bernoulli}(\pi_k)  \label{bern_prior} 
\eeqs 

We further assume a Gaussian prior on membership strengths $\rv_n \in \mathbb{R}^K$, i.e., $p(\rv_n) = \Ncal(\mathbf{0},\sigma^2\Imat)$.


\subsection{VAE Encoder}

We employ a \emph{nonlinear} encoder to infer the node embedding $\zv_n$ for each node, using a fast non-iterative \emph{recognition model}~\cite{kingma2013auto}. Denoting the parameters of the variational posterior for the embeddings of all the nodes collectively as $\{\vv, \bv, \rv\}$, we consider an approximation to the model's true posterior $p(\vv,\bv,\rv|\Amat, \Xmat)$ with a variational posterior of the form $q_\phi(\vv,\bv,\rv)$. For simplicity, we consider a mean-field approximation, which allows us to factorize the posterior as $q_\phi(\vv,\bv,\rv) = \prod_{k=1}^K \prod_{n=1}^{N} q_\phi(v_{nk})q_\phi(b_{n,k})q_\phi(r_{n,k})$.
Our nonlinear encoder, as shown in  Fig. \ref{fig:dec-enc} (Right), assumes variational distributions on the local variables of each node, $i.e.$, $\vv_n$, $\bv_n$ and $\rv_n$, and defines the variational parameters of these distributions as the outputs of a graph convolutional network (GCN)~\cite{kipf2016semi}, which takes as input the network $\Amat$ and the node feature matrix $\Xmat$. GCN has recently emerged as a flexible encoder of graph-structured data (similar in spirit to convolutional neural networks for images), which makes it an ideal choice of the encoder in our VAE-based generative model for graphs. The forward propagation rule for each layer $l$ in GCN is defined as $\Hmat^{l} = g(\hat{\Amat}\Hmat^{l-1}\Wmat^{l})$, where $\Hmat^{0} = \Xmat$ ($\Xmat = \Imat$ when no side information is present), $\Wmat^{l}$ is the weight matrix, $g(\cdot)$ is the non-linear activation, and $\hat{\Amat}$ is the symmetric normalization of adjacency $\Amat$. Although here we have used the vanilla GCN in our architecture, more-generalized variants of GCN, such as GraphSAGE~\cite{hamilton2017inductive}, can also be used as the encoder. The variational distributions have the following forms
\beqs
  q_\phi(v_{nk}) &=& \text{Beta}(c_{nk}, d_{nk}) \quad k=1,\ldots,K  \label{beta_post} \\ 
  q_\phi(b_{nk}) &=& \text{Bernoulli}(\pi_{nk}) \quad k=1,\ldots,K \label{bern_post} \\
  q_\phi(\rv_n) &=& \Ncal(\boldsymbol{\mu}_n,\text{diag}(\boldsymbol{\sigma}_n^2)) 
\eeqs
where $c_{nk}, d_{nk}, \pi_{nk}, \muv_n, \text{and } \sigmav_n$ are outputs of a GCN, $i.e.$, $\{c_k, d_k, \pi_k, \mu_k, \sigma_k\}_{n=1}^{n=N} = \text{GCN}(\Amat, \Xmat)$. We use the stochastic gradient variational Bayes (SGVB) algorithm~\cite{kingma2013auto} to infer the parameters of the variational distributions. Details on reparameterization and the loss formulation are provided in Section \ref{inference}.

\subsection{Special Cases}

Existing models for graph-structured data can be seen as special cases of our framework. Recall that we model the node embeddings as $\zv_n = \bv_n \odot \rv_n$, and our generative model is of the form $A_{nm} \sim p_\theta(\zv_n,\zv_m)$. If we ignore the community strength latent variable $\rv_n$, $i.e.$, $\zv_n$ is defined simply as $\zv_n = \bv_n$ (just a binary vector) and further define $p_\theta$ as a Bernoulli distribution with its probability being a bilinear function of the embeddings $\zv_n$ and $\zv_m$, then we recover the OSBM/LFRM~\cite{latouche2011overlapping,miller2009nonparametric}. Note, however, that while OSBM/LFRM typically rely on MCMC or variational inference, our framework can leverage SGVB for efficient inference.

Likewise, if we define $\zv_n = \rv_n$, $i.e.$, a \emph{dense} vector, and define $p_\theta$ as a Bernoulli distribution with its probability being a bilinear function of the embeddings, we recover the Eigenmodel or latent-space model (LSM)~\cite{hoff2002latent}. Note that this model cannot infer $K$ since the binary vector $\bv_n$ is not present. Finally, if $p_\theta$ is a Bernoulli distribution with its probability being a \emph{nonlinear} function of the embeddings, then we recover the VGAE model ~\cite{kipf2016variational}, which can also be seen as a nonlinear extension of LSM. Moreover, note that a key limitation of LSM and VGAE is that these cannot be used to infer the community structure (due to the non-sparse nature of $\zv_n$) and usually can only be used for link-prediction tasks.

\vspace{-1em}
\section{Inference}
\label{inference}
We define the factorized variational posterior $q_\phi(\vv,\bv,\rv)$ as
\beqs
q_\phi(v_{nk})        &=& \text{Beta}(v_{nk}|c_k(\Amat,\Xmat),d_k(\Amat,\Xmat)) \nonumber \\
q_\phi(b_{nk}) &=& \text{Bernoulli}(b_{nk}|\pi_{k}(\Amat,\Xmat)) \nonumber \\
q_\phi(\rv_n) &=& \Ncal(\boldsymbol{\mu}_n(\Amat,\Xmat),\text{diag}(\boldsymbol{\sigma}_n^2(\Amat,\Xmat))) \nonumber
\eeqs
where $c_k, d_k, \pi_k, \boldsymbol{\mu}_n \text{and } \boldsymbol{\sigma}_n$ are a function of the GCN encoder, with inputs $\Amat$ and $\Xmat$. We define the loss function $\mathcal{L}$ parameterized by inference network (encoder) parameters ($\phi$) and generator parameters ($\theta$) by minimizing the \emph{negative} of the evidence lower bound (ELBO)
\beq
\small
\medmuskip=2mu 
\begin{aligned}
	\label{eqn:loss}
    \mathcal{L} =& \sum_{n=1}^N \Big(\mathrm{KL}[q_\phi(\bv_n|\boldsymbol{v}_n)\,||\,p_\theta(\bv_n|\boldsymbol{v}_n)] + \mathrm{KL}[q_\phi(\rv_n)\,||\,p_\theta(\rv_n)] \\ &+ \mathrm{KL}[q_\phi(\boldsymbol{v}_n)\,||\,p(\boldsymbol{v}_n)]\Big)
    - \sum_{n=1}^N \Big( \mathbb{E}_{q}[\log p_\theta(X_{n}|\zv_n)]\Big) \\
&- \sum_{n=1}^N \sum_{m=1}^N \Big( \mathbb{E}_{q}[\log p_\theta(A_{nm}|\zv_n,\zv_m)]\Big)
\end{aligned}
\normalsize
\eeq

where $\mathrm{KL}[q(\cdot)||p(\cdot)]$ is the Kullback-Leibler divergence between $q(\cdot)$ and $p(\cdot)$. Note that here we have also included the loss from the reconstruction of the side information $X_n$. We have considered that the side information $\Xmat$ and the links $\Amat$ are conditionally independent given the node embeddings $\zv_{1:N}$. When there is no side information, we can ignore the $\mathbb{E}_{q}[\log p_\theta(X_{n}|\zv_n)]$ term in the loss function. For the encoder and decoder parameters we infer point estimates, while we learn the distribution over the latent variables $\bv, \vv, \text{ and } \rv$.

Our variational autoencoder for link generation is trained using Stochastic Gradient Variational Bayes (SGVB) \cite{kingma2013auto}. SGVB can be used to perform inference for a broad class of non-conjugate models and is therefore appealing to Bayesian nonparametric models, such as those based on stick-breaking priors that we use in our framework. SGVB uses differentiable Monte Carlo (MC) expectations to learn the model parameters. Specifically, it requires \emph{differentiable, non-centered parameterization} (DNCP) \cite{KingmaW14} to allow backpropagation. However, our model has expectations over Beta and Bernoulli distributions, neither of which permit easy reparameterization as required by SGVB. We leverage recent developments on reparameterizing these distributions~\cite{MaddisonMT16,ErickSBVAE}, which consequently leads to a simple inference procedure.

Following \cite{ErickSBVAE}, we approximate the Beta posterior in (\ref{beta_post}) with the Kumaraswamy distribution, defined as: $Kumar(x;a,b) = abx^{a-1}(1-x^a)^{b-1}$ for $x \in (0,1)$ and $a,b > 0$. The closed-form inverse CDF allows easy reparameterization, and samples for $v_{nk}$ (with parameters $c_{nk}$ and $d_{nk}$) can be drawn using: 
\beqs
u &\sim& \text{Uniform}(0,1) \nonumber \\ 
v_{nk} &\stackrel{d}{=}& (1-u^{\frac{1}{d_{nk}}})^{\frac{1}{c_{nk}}}
\eeqs
We compute the KL divergence between the Kumaraswamy $q(\vv)$ and the Beta distribution $p(\vv)$ by taking a finite approximation of the infinite sum as mentioned in \cite{ErickSBVAE}. 

For the Bernoulli random variable, we use the Binary Concrete distribution \cite{MaddisonMT16,jang-categorical-gumbel} at the time of training, as a continuous relaxation to get the biased low-variance estimates of the gradient. The $\mathrm{KL}$ divergence between two Bernoulli distributions is relaxed using two Binary Concrete distributions.

We reparameterize $b_{nk}$, defined by a Bernoulli with probability $\pi_{nk}$, (in (\ref{bern_prior}) and (\ref{bern_post})) with reparameterization:
\beqs
L &=& \log\Big(\frac{u}{1-u}\Big) \nonumber \\
b_{nk} &\stackrel{d}{=}& \sigma\,\Bigg( \frac{\text{logit}(\pi_{nk}) + L}{\lambda} \Bigg)
\eeqs
where $\sigma(\cdot)$ is the sigmoid function, $logit(\cdot)$ is the inverse-sigmoid function, $\lambda$ is the relaxation temperature and $u \sim \text{Uniform}(0,1)$.

\textbf{Structured Mean-Field:} Since the vanilla mean-field variational inference ignores the posterior dependence among the latent variables, we also considered Structured Stochastic Variational Inference (SSVI) \cite{DBLP:journals/corr/Hoffman14,Hoffman:2013:SVI:2502581.2502622}, which allows global-local parameter dependency and improves upon the mean-field approximation. We considered $\vv$ (and its variational parameters $c \text{ and } d$) as global parameters and impose a hierarchical structure on $\bv_n$ by conditioning it on $\vv$. The variational posterior of our framework using SSVI can be factorized as $q_\phi(\vv,\bv,\rv) = \prod_{k=1}^Kq_\phi(v_{k})\prod_{n=1}^{N}q_\phi(b_{n,k}|\vv)q_\phi(r_{n,k})$ with $q_\phi(v_{k}) = \text{Beta}(c_{k}, d_{k}); q_\phi(b_{nk}|\vv) = \text{Bernoulli}(\pi_{k}); \pi_k = \prod_{j=1}^K v_k$, where $c_k, d_k$ are parameters to be learned. In practice, we have found structured mean-field to perform better than the mean-field, and our model implementation uses the former.

\section{Related Work}
The proposed framework can be seen as bridging two lines of research on modeling graphs: ($i$) structured latent variable models for graphs, such as stochastic blockmodels and its variants~\cite{kemp2006learning,airoldi2008mixed,miller2009nonparametric,latouche2011overlapping}; and ($ii$) deep learning models for graphs, such as graph convolutional networks~\cite{kipf2016semi}. Our effort is motivated by the goal of harnessing their complementary strengths to develop a deep generative stochastic blockmodel for graphs, that also enjoys an efficient inference procedure. 

The most prominent methods in stochastic blockmodels include models that associate each node to a single community~\cite{nowicki2001estimation,kemp2006learning}, a mixture of communities~\cite{airoldi2008mixed}, and an overlapping set of communities~\cite{miller2009nonparametric,latouche2011overlapping,yang2012community,zhou2015infinite}. While stochastic blockmodels have nice interpretability, these models usually assume the links of the networks to be modeled as a simple bilinear function of the node embeddings, which may not be able to capture the nonlinear interactions between the nodes~\cite{yan2011sparse}. An approach to model such nonlinear interactions was proposed in~\cite{yan2011sparse}, using a matrix-variate Gaussian process. However, despite the modeling flexibility, inference in this model is challenging and the model is usually infeasible to run on networks with more than 100 nodes. 

There is also significant recent interest in non-probabilistic deep learning models for graphs. Some of the prominent works in this direction include DeepWalk~\cite{perozzi2014deepwalk} and graph autoencoders (GAE)~\cite{kipf2016semi,hamilton2017inductive}. DeepWalk is inspired by the idea of word embeddings. It treats each node as a ``document,'' by starting a random walk at that node and taking the nodes encountered in the path taken as the word in that document. It uses document/word embedding methods to the learn embedding of each node. In contrast, the GAE approaches are based on the idea of graph convolutional networks (GCN)~\cite{kipf2016semi}. This line of work nicely complements our contribution, since modules like GCN can be effectively used to design the encoder model for our deep generative framework. In particular, as noted in the model description, our encoder is essentially a GCN. We believe that such advances in graph encoding can be used as modules to design new deep generative models for relational data. 

Despite the significant success of deep generative models for images and text data, there has been relatively little work on deep generative models for relational data~\cite{graphrnn,hu2017deep,wang2017relational,kipf2016variational}. GraphRNN~\cite{graphrnn} learns a single representation of an entire graph to model the joint distribution of different graphs. The focus of GraphRNN is on generating small-sized graphs, whereas we focus on link prediction and community detection for a given graph. Among other existing methods, ~\cite{hu2017deep} proposed an extension of the LFRM via a deep hierarchy of binary latent features for each node. However, this model relies on expensive batch MCMC inference, precluding its applicability to large-scale networks. Another deep latent variable model was proposed recently in~\cite{wang2017relational}. However, this model also has a difficult inference procedure, requiring model-specific inference. Moreover, the node embeddings are not interpretable. Perhaps the closest in spirit to our work is the recent work on variational graph autoencoders (VGAE)~\cite{kipf2016variational}. Graphite~\cite{grover2018graphite} extends the VGAE by using a multi-layer iterative decoder that alternates between message passing and graph refinement. A similar decoding scheme can also be applied in our framework; however, the focus of this work is on learning sparse interpretable node embeddings. Both VGAE and Graphite are built on top of the standard VAE, and consequently do not have direct interpretability of node embeddings as desired by stochastic blockmodels. This leads to a model with different properties and a different inference procedure, compared to~\cite{kipf2016variational}. Moreover, our VAE architecture is nonparametric in nature and can infer the node embedding size. 

\section{Experiments}

We report experimental results on several synthetic and real-world datasets, to demonstrate the efficacy of our model. Our experimental results include quantitative comparisons on the task of link prediction as well as qualitative results, such as using the embeddings to discover the underlying communities in the network data. The qualitative results are meant to demonstrate the expressiveness of the latent space that our model infers. The expressive nature of our model is the result of the sparse and interpretable embedding for each node of the graph. In particular, we show that these sparse embeddings can be interpreted as the memberships and strength of memberships of each node in one or more communities.

First we evaluate our model on link-prediction, comparing it with various baselines on several benchmark datasets on moderate (about 2000 nodes) to large-scale (about 20,000 nodes) datasets. We then analyze the latent structure $\zv_n$ learned by our model on a synthetic and a real-world co-authorship dataset. We compare the latent structure with the embeddings learned by the variational graph autoencoder (VGAE)~\cite{kipf2016variational}. We also examine the community structure on the real-world co-authorship dataset, and show that the proposed framework is able to readily capture the underlying communities. We refer to our framework as DGLFRM (Deep Generative Latent Feature Relational Model), which refers to our most general model with sparse embeddings $\zv_n = \bv_n \odot \rv_n$ with nonlinear generator and nonlinear encoder. We also consider a variant of DGLFRM with binary embeddings $\zv_n = \bv_n$, which we refer to as DGLFRM-B (the `B' here denotes ``binary''). Note that DGLFRM-B can be seen as a deep generalization of LFRM~\cite{Miller:2009:NLF:2984093.2984237}/OSBM~\cite{latouche2011}, with another key difference from LFRM/OSBM being the fact that we use amortized inference.  

\subsection{Baselines}

For link prediction, we compare the proposed model with four baselines, one of which is a simplified variant of DGLFRM akin to LFRM~\cite{miller2009nonparametric}, which is an overlapping stochastic blockmodel. The original LFRM, which uses MCMC-based inference, was infeasible to run on the datasets used in these experiments. On the other hand, DGLFRM with $\zv_n = \bv_n$ and bilinear decoder (link generation model) is similar to LFRM, but with a much faster SGVB based inference (we will refer to this simplified variant of DGLFRM as LFRM).  

Among the other three baselines, Spectral Clustering (SC) and DeepWalk (DW)~\cite{perozzi2014deepwalk} learn node embeddings, which we use to compute the link probability as $\sigma(\zv_n^\top \zv_m)$. The third baseline is the recently proposed variational autoencoder on graphs (VGAE)~\cite{kipf2016variational}. Note that none of these baselines can be used for community detection, since the real-valued embeddings learned by these baselines are not interpretable (unlike our model which learns sparse embeddings, with nonzeros denoting community memberships).

\begin{table*}[!ht]
\footnotesize
  \caption{AUC ROC}
\vspace{0.5em}
  \centering
  \begin{tabular}{l c c c c c}
    \toprule
    \textbf{Method}   & \textbf{NIPS12} & \textbf{Yeast} & \textbf{Cora}     & \textbf{Citeseer} & \textbf{Pubmed} \\
    \midrule
    
    SC & $0.8792\pm.0003$ & $0.7886\pm.0001$ & $0.8460\pm.0001$& $0.8050\pm.0001$ & $0.8420\pm.0002$  \\
    DW  & $0.8058\pm.0000$ & $0.6443\pm.0003$ & $0.8310\pm.0001$ & $0.8050\pm.0002$ & $0.8440\pm.0000$  \\
    VGAE & $0.8790\pm.0055$ & $0.7784\pm.0002$ & $0.9260\pm.0001$ & $0.9080\pm.0002$ & $\textbf{0.9418}\pm.0076$  \\
    LFRM      			& $0.8489 \pm .0001$ & $0.7975\pm.0006$ &$0.9096\pm.0026$&$0.8965\pm.0035$ & $0.9152\pm.0041$ \\
    \midrule
    DGLFRM-B				& $\textbf{0.8898}\pm.0028$ & $\textbf{0.8061}\pm.0003$ & $0.9281\pm.0024$ & $0.9007\pm.0020$ & $0.9396\pm.0052$ \\
DGLFRM     			& $0.8734\pm.0043$ & $0.7856\pm.0005$ & $\textbf{0.9343}\pm.0023$& $\textbf{0.9379}\pm.0032$ & $0.9395\pm.0008$ \\
    \bottomrule
  \end{tabular}
  \normalsize
  \label{table:roc_scores}
\end{table*}

\begin{table*}[!ht]
\footnotesize
  \caption{Average Precision (AP).}
\vspace{0.5em}
  \centering
  \begin{tabular}{l c c c c c}
    \toprule
    \textbf{Method}  & \textbf{NIPS12} & \textbf{Yeast} & \textbf{Cora}     & \textbf{Citeseer} & \textbf{Pubmed} \\
    \midrule
    SC  & $0.9022\pm.0002$ & $0.8440\pm.0001$ & $0.8850\pm.0000$ & $0.8500\pm.0100$ & $0.8780\pm.0100$  \\
    DW  & $0.8634\pm.0000$ & $0.6699\pm.0002$ & $0.8500\pm.0001$ & $0.8360\pm.0001$ & $0.8440\pm.0000$  \\
    VGAE & $0.9114\pm.0042$ & $0.8349\pm.0002$ & $0.9328\pm.0001$ & $0.9200\pm.0002$ & $0.9394\pm.0088$  \\
    LFRM      			& $0.8870\pm.0000$ & $0.8268\pm.0005$ &$0.9060\pm.0033$&$0.9118\pm.0031$ & $0.9197\pm.0054$ \\
    \midrule
    DGLFRM-B				& $\textbf{0.9120}\pm.0021$ & $\textbf{0.8442}\pm.0002$ & $0.9259\pm.0023$ & $0.9153\pm.0031$ & $0.9454\pm.0050$    \\
    DGLFRM     			& $0.9005\pm.0027$ & $0.8388\pm.0002$ & $\textbf{0.9376}\pm.0022$& $\textbf{0.9438}\pm.0073$ & $\textbf{0.9497}\pm.0035$\\
    \bottomrule
  \end{tabular}
  \normalsize
  \label{table:ap_scores}
\end{table*}

\begin{table*}[!ht]
\begin{center}
\footnotesize
\caption{Example of communities inferred by our model on the NIPS data.}
\label{sample-table}
\vspace{0.5em}
\begin{tabular}{ll}
\toprule
Cluster & Authors\\
\midrule
Probabilistic Modeling & Sejnowski T, Hinton G, Dayan P, Jordan M, Williams C \\
Reinforcement Learning & Barto A, Singh S, Sutton R, Connolly C, Precup D \\
Robotics/Vision & Shibata T, Peper F, Thrun S, Giles C, Michel A \\
Computational Neuroscience & Baldi P, Stein C, Rinott Y, Weinshall D, Druzinsky R \\
Neural Networks & Pearlmutter B Abu-Mostafa Y, LeCun Y, Sejnowski T, Tang A\\
\bottomrule
\end{tabular}
\end{center}
\end{table*}

\subsection{Datasets}

We consider five real-world datasets, with three datasets consisting of side information in the form of the node features, and the other two datasets having only the link information. For the link-prediction experiments, all models are provided a partially-complete network (with unknown part to be predicted). The node features (when available) are provided to all the models. The description of each data set is as follows:

\begin{itemize}
\item \textbf{NIPS12}: The NIPS12 coauthor network \cite{zhou2015infinite} includes all 2037 authors in NIPS papers from volumes 1-12, with 3134 edges. This network has no side information.
\item \textbf{Yeast}: The Yeast protein interaction network \cite{zhou2015infinite} has 2361 nodes and 6646 non-self edges. This network has no side information.
\item \textbf{Cora}: Cora network is a citation network consisting of 2708 documents. The datasets contain sparse bag-of-words feature vectors of length 1433 for each document. These are used as node features. The network has total 5278 links.
\item \textbf{Citeseer}: Citeseer is a citation network consisting of 3312 scientific publications from six categories: agents, AI, databases, human computer interaction, machine learning, and information retrieval. The side information for the dataset is the category label for each paper which is converted into a one-hot representation. These one-hot vectors are used as node features. The network has a total of 4552 links.
\item\textbf{Pubmed}: A citation network consisting of 19,717 nodes. The dataset contains sparse bag-of-words feature vectors of length 500 for each document, used as node features. The network has total 44,324 links.
\end{itemize}

\subsection{Link Prediction}

We use Area Under the ROC Curve (AUC) and Average Precision (AP) to compare our model with the other baselines for link prediction. For all datasets, we hold out 10\% and 5\% of the links as our test set and validation set, respectively, and use the validation set to fine-tune the hyperparameters. We take the average of AUC-ROC and AP scores by running our model on 10 random splits of each dataset, to compare with the baselines. The AUC-ROC scores of our models and the various baselines are shown in Table \ref{table:roc_scores} and AP scores are shown in Table \ref{table:ap_scores}. As shown in the tables, our models outperforms the baselines on almost all datasets. We again highlight that unlike the baselines, such as VGAE, that cannot learn interpretable embeddings, our model also learns embeddings that can be interpreted as memberships of nodes into communities. The superior results of DGLFRM and DGLRFM-B demonstrate the benefit of our deep generative models. The significantly better results of these as compared to LFRM also show the benefit of endowing LFRM with a deep architecture, with nonlinear decoder and nonlinear encoder. The hyperparameter settings used for all experiments are included in the Supplementary Material. We also performed an experiment to investigate the model's ability to leverage node features. As expected, when using the features the model performs better compared to the case when it ignores features. This experiment is included in the supplementary section.

\begin{figure*}[!ht]
 \centering
\subfigure[Adjacency]{\includegraphics[width=0.24\textwidth]{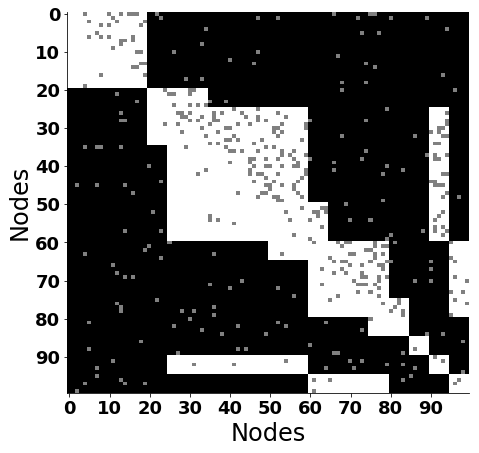}}
\hfill
\subfigure[DGLFRM]{\includegraphics[width=0.24\textwidth]{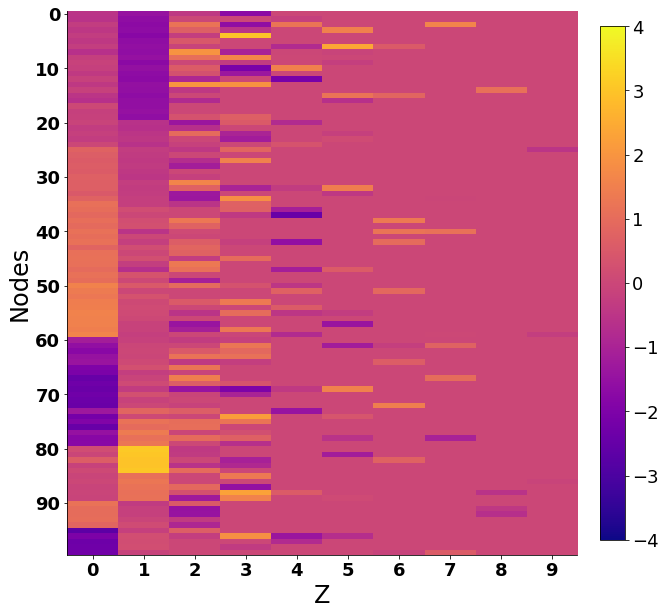}}
\hfill
\subfigure[Generated]{\includegraphics[width=0.265\textwidth]{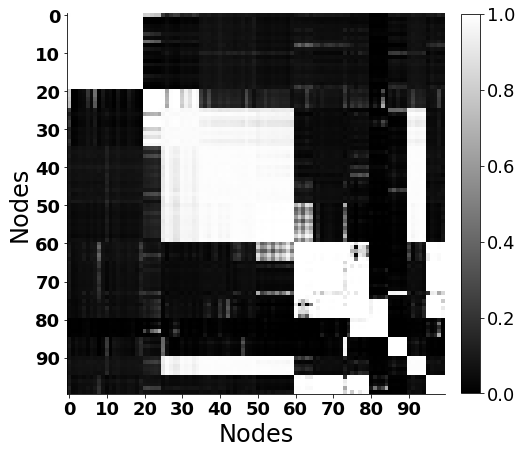}}
\hfill
\subfigure[VGAE]{\includegraphics[width=0.24\textwidth]{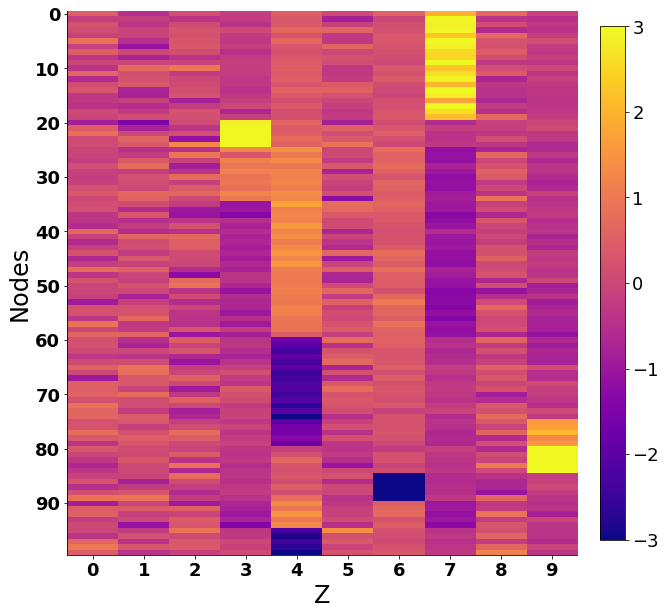}}
\caption{\small{(a) The synthetic adjacency matrix (where white, black and grey denote link, no-link and hidden parts of the graph). (b) The latent structure of synthetic data learned using DGLFRM. The truncation parameter (K) of the latent structure was fixed to 10. (c) The graph generated by DGLFRM using only the first 2 dimensions of the latent structure($i.e.$ columns 0 and 1 in (b)). The columns 2-9 were all set to zero. The graph is represented as the probability of links; white (black) represents link (no-link) with high probability. (d) The latent structure of synthetic data learned using VGAE.}} 
\label{fig:qual}
\end{figure*}

\subsection{Qualitative Analysis on Learned Embeddings}
\label{sec:qualitative}

To demonstrate the interpretable nature of the embeddings learned by our model, we generate a synthetic dataset with 100 nodes and 10 communities. The dataset is generated by fixing the ground-truth communities (by creating a binary vector for each node) such that some of the nodes belong to same communities. The adjacency matrix is then generated using a simple inner product, followed by the sigmoid operation (Figure \ref{fig:qual}a). We train using 85\% of the synthetic adjacency matrix for link-prediction and for visualizing the latent structure that our model learns. The latent structure obtained using DGLFRM is plotted in Figure \ref{fig:qual}(b). Figure \ref{fig:qual}(c) shows that by using only the first two dimensions of the latent structure we can reconstruct the graph reasonably well. This depicts an important property of using a stick-breaking IBP prior which encourages the most commonly selected communities (the columns on the left in Figure \ref{fig:qual} (b)) to be dense, while the communities with higher indices (columns in right) to be sparse. This shows that DGLFRM can learn the effective number of communities given a graph. Finally, we can quantize the latent space into discrete intervals to extract nodes belonging to different communities. In our experiments we saw that the latent structure learned is in fact close to the ground-truth community assignments we started with. In Figure \ref{fig:qual}(d) we compare the community structure from our model with the latent structure obtained by running the VGAE. Note that the Gaussian latent structure in VGAE is dense and therefore fails to learn community memberships that are readily interpretable.

We also do a qualitative analysis on the NIPS12 dataset. Again we train DGLFRM and VGAE using 85\% of the adjacency matrix. Table~\ref{sample-table} shows five of the inferred communities by DGLFRM. The authors shown under each community are ordered by the strength of their community memberships (in decreasing order). As Table~\ref{sample-table} shows, each of the communities represent a sub-field, with authors working on similar topics. Moreover, note that some authors ($e.g.$, Sejnowski) are inferred as belonging to more than one community. This qualitative experiment demonstrates that our model can learn interpretable embeddings that can be used for tasks such as (overlapping) clustering. We have included a visualization of the latent structure learned on NIPS12 data in the Supplementary Material. Note that our model can infer the number of communities naturally, via the stick-breaking prior. The stick-breaking prior requires specifying a large truncation level on the number of communities. Our model effectively infers the ``active'' communities for a given truncation level. As shown in Fig.~\ref{fig:qual} (b)-(c), the posterior inference in our model is able to ``turn off'' the unnecessary columns in $\Zmat$. Although we do not know the ground truth for the number of communities, the number of inferred active communities is similar to what is reported in prior work on nonparametric Bayesian overlapping stochastic blockmodels~\cite{miller2009nonparametric}. Note that VGAE embeddings require an additional step (such as $K$-Means clustering) to cluster nodes. Moreover, a method such as $K$-means cannot detect overlapping communities, and it is also sensitive to the initialization of $K$ (estimated number of communities). For reference, we have included the clustering results on the VGAE embeddings in the Supplementary Material.  

\section{Conclusion and Discussion}
We have presented a deep generative framework for overlapping community discovery and link prediction. This work combines the interpretability of stochastic blockmodels, such as the latent feature relational model, with the modeling power of deep generative models. Moreover, leveraging a nonparametric Bayesian prior on the node embeddings enables learning the node embedding size ($i.e.$, the number of communities) from data. Our framework is modular and a wide variety of decoder and encoder models can be used. In particular, it can leverage recent advances in non-probabilistic autoencoders for graphs, such as the graph convolutional network~\cite{kipf2016semi} or its extensions~\cite{hamilton2017inductive}. Inference in the model is based on SGVB, that does not require conjugacy. This further widens the applicability of our framework to model different types of networks ($e.g.$, weighted, count-valued edges, and power-law degree distribution of node degrees). We believe this combination of discrete latent variables based stochastic blockmodels and graph neural network will help leverage their respective strengths, and will fuel further research and advance the state-of-the-art in (deep) generative modeling of graph-structured data.

Although SGVB inference makes our model fairly efficient, it can be scaled up further for massive networks by using mini-batch based inference \cite{chen2018iclr}. Another possibility to scale up the model is to replace the Bernoulli-logistic likelihood model by a Bernoulli-Poisson link~\cite{zhou2015infinite}, which enable scaling up the model in the number of nonzeros ($i.e.$, number of edges) in the network. Given that our framework can work with a wide variety of decoder/generator models, such modifications can be done without much difficulty.

Finally, in this work we model each node as having a single binary vector, denoting its memberships in one or more communities. Another interesting extension would be to consider multiple layers of latent variables, which can model a node's membership into a hierarchy of communities~\cite{ho2011multiscale,blundell2013bayesian,hu2017deep}.

\small
\textbf{Acknowledgements:} PR acknowledges support from Google AI/ML faculty award and DST-SERB Early
Career Research Award. The Duke investigators acknowledge the support of DARPA and ONR.
\normalsize

\clearpage
\small
\bibliography{paper}
\bibliographystyle{icml2019}

\section{Supplementary Material}
\subsection{Hyparameter Settings}
\subsubsection*{Quantitative results:}

The framework proposed uses a stick-breaking IBP prior which has two parameters: $\alpha$ and $K$. The parameter $\alpha$ is the initial guess of the number of non-zero entries in the binary vector $b_n$ and $K$ is the truncation parameter. In the experiments, $\alpha \in \{5, 10, 20, 50, 100\}$. In general, a higher value of the $\alpha$ parameter worked better for DGLFRM-B and LFRM, as compared to the $\alpha$ value in the DGLFRM model. This difference in $\alpha$ reflects the inherent capacity of the latent space of these models. The embedding learned by DGLFRM, while being highly sparse, are in real space resulting in more capacity to represent data as compared to the binary latent space in DGLFRM-B and LFRM.  

The encoder network for DGLFRM and DGLFRM-B had two non-linear GCN layers. The length of the first non-linear layer was fixed to 32/64 for the datasets which had side-information (Cora, Citeseer and Pubmed), and otherwise was set to 128/256. The second layer of the GCN encoder had $K \in \{50, 100, 200\}$ hidden units. The decoder network for DGLFRM and DGLFRM-B had two layers with dimension 32 and 16. All the models were trained for 500-1000 iterations using the Adam optimizer \cite{AdamKingmaB14} with a learning rate of $0.01$. We used 0.5 dropout. The temperature parameter of the Binary Concrete distribution~\cite{MaddisonMT16} was 0.5 for the prior and 1.0 for the posterior. 

\subsubsection*{Qualitative results:}
For experiments on the synthetic data (with 100 nodes and 10 communities), the DGLFRM model had two GCN encoding layers with 32 and $K=10$ hidden units, and the decoder had a simple inner-product layer. The VGAE model had the same set of hyperparameters as above. The qualitative experiment on the NIPS12 co-authorship dataset had two hidden layers with 64 and $K=10$ hidden units. The $\alpha$ parameter for this experiment was fixed to 2.

\subsection{K-means on VGAE embeddings}
Variational Graph Autoencoder (VGAE), unlike the proposed model, is not able to learn embeddings which are readily interpretable. It requires additional processing such as K-Means over the learned embeddings for node clustering. Moreover, a method like K-Means does not result in overlapping communities. In this section, we compare the clusters obtained after applying K-means on embeddings learned from VGAE with the readily available overlapping communities obtained from our framework.

We use K-means to find clusters for NIPS12 (3134 authors) co-authorship data on the node embeddings learned using VGAE. The K-means results are shown in Table~\ref{kmeans}. We performed two experiments with different k-means cluster hyperparameter $K$ (K=5 and K=20). We also show the clusters (communities) which our model was readily able to infer for reference Table~\ref{sample-table}. We only show prominent authors and their clusters for both the models.

As we see in Table \ref{kmeans}, ad-hoc post-processing of embeddings may break some relevant coherent communities which were inferred by our model. It is also important to note that, unlike our model, k-means has no strength indicator for community membership.

\subsection{Latent Structure on NIPS12}

The latent structure of the NIPS12 dataset learned by DGLFRM and VGAE is shown in Figure \ref{fig:qual}. In this experiment, the truncation parameter for the stick-breaking prior is 50. As shown in Figure \ref{fig:qual}(b), the posterior inference in DGLFRM is naturally able to ``turn off'' the unnecessary columns in $\Zmat$. The average number of active communities for each node was found to be 8. The sparse nature of the embedding matrix allows us to consider each column as a possible community of a given node. For visualization, we have ordered the indices of the communities (columns of $\Zmat$) such that the community with higher active nodes has a lower index in the visualization. For the VGAE model, we used a two layer GCN with dimensions 32 and 16. Figure \ref{fig:qual}(c) depicts the dense node embedding learned by VGAE. 

\subsection{Effect of Side Information}

\begin{figure*}[!ht]
 \centering
\subfigure[DGLFRM]{\includegraphics[width=0.32\textwidth]{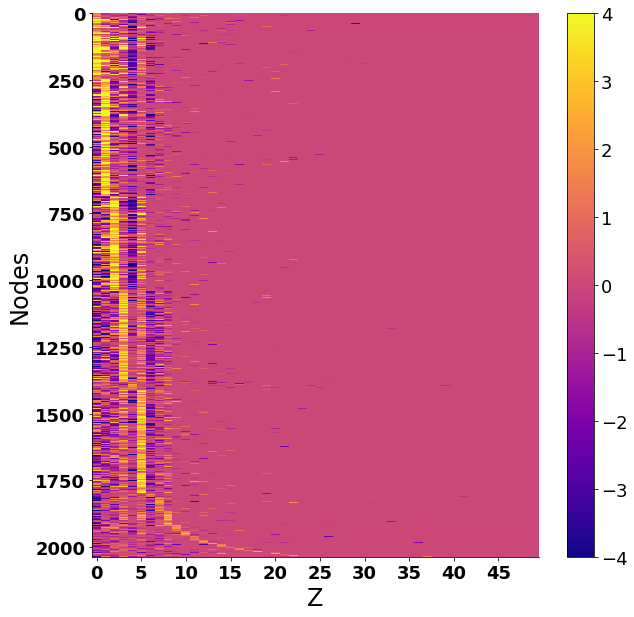}}
\hfill
\subfigure[DGLFRM (Filtered)]{\includegraphics[width=0.32\textwidth]{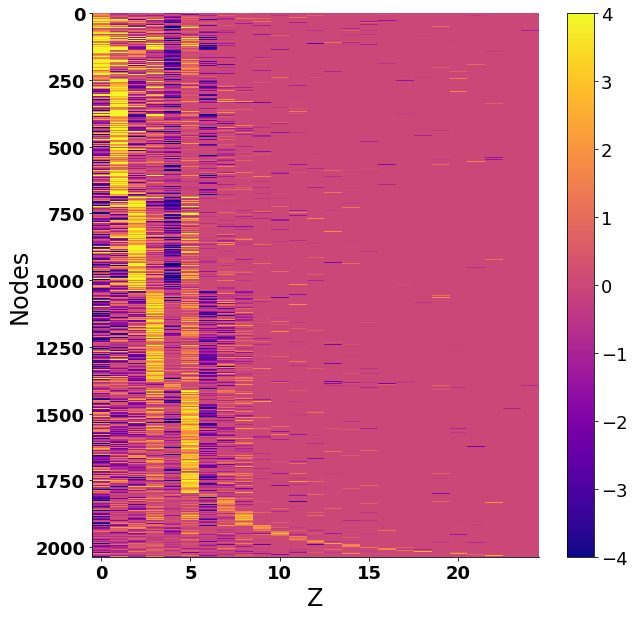}}
\hfill
\subfigure[VGAE]{\includegraphics[width=0.32\textwidth]{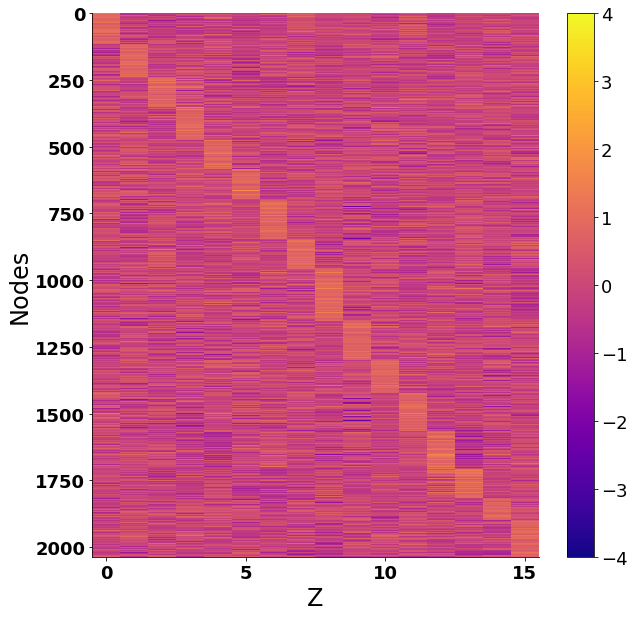}}
\caption{\small{(a-c) The latent structure on NIPS12 dataset using DGLFRM and VGAE. Both the models had the same encoder and latent dimension. The latent structure learned by DGLFRM was filtered by removing the columns which were inactive for all nodes. DGLFRM can effectively infer the ``active'' communities.}}
\label{sideinfo}
\end{figure*}

We also perform an experiment to investigate the model's ability to leverage side information associated with nodes. For this experiment, we ran our model on three datasets (Cora, Citeseer and Pubmed) with and without node features. We compare the AUC-ROC results in Fig.~\ref{sideinfo}. As expected, when using the node side information, the model performs better as compared to the case when it ignores the side information.

\begin{figure}[!h]
\begin{center}
\centerline{\includegraphics[scale=0.3]{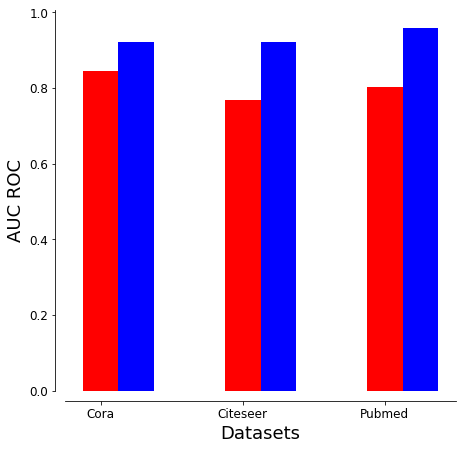}}
\vspace{-1em}
\caption{\small{Red: Without side information. Blue: With side information.}}
\end{center}
\end{figure}

\begin{table*}[!htbp]
\caption{Example of NIPS12 communities inferred by k-means clustering (post-processing step) on the embeddings learned using VGAE. Authors have hard-assignments (memberships) in these communities.}
\vskip 0.15in
\begin{center}
\begin{small}
\begin{tabular}{ll}
\toprule
Cluster (K=5) & Authors \\
\midrule
Cluster 1 & Hinton G, Dayan P, Jordan M, Tang A, Sejnowski T, Willams C \\
Cluster 2 & Weinshall D, Rinott Y, Barto A, Singh S, Sutton R, Giles C, Connolly C, Baldi P, Precup D \\
Cluster 3 & Thrun S, Shibata T, Stein C, Peper F, Michel A, Druzinsky R, Abu-Mostafa Y \\
Cluster 4 & LeCun Y, Pearlmutter B \\
\midrule
Cluster (K=20) & Authors \\
\midrule
Cluster 1 & Hinton G, Williams C \\
Cluster 2 & Jordan M, Connolly C, Barto A, Singh S, Sutton R\\
Cluster 3 & Michel A, Tang A \\
Cluster 4 & Dayan P, Sejnowski T\\
Cluster 5 & Thrun S, Peper F\\
Cluster 6 & Baldi P, Weinshall D \\
Cluster 7 & Shibata T, Druzinsky R\\
Cluster 8 & Stein C\\
Cluster 9 & Precup D\\
Cluster 10 & Giles C\\
Cluster 11 & Pearlmutter B\\
Cluster 12 & LeCun Y\\
Cluster 13 & Rinott Y\\
Cluster 14 & Abu-Mostafa Y\\
\bottomrule
\end{tabular}
\end{small}
\end{center}
\label{kmeans}
\vskip 0.15in
\end{table*}

\begin{table*}[!ht]
\caption{Example of communities inferred by our model on NIPS data. Authors ordered by  strength of membership in these communities.}
\vskip 0.15in
\begin{center}
\small
\begin{tabular}{ll}
\toprule
Cluster & Authors\\
\midrule
Probabilistic Modeling & \textbf{Sejnowski T}, Hinton G, Dayan P, Jordan M, Williams C \\
Reinforcement Learning & Barto A, Singh S, Sutton R, Connolly C, Precup D \\
Robotics/Vision & Shibata T, Peper F, Thrun S, Giles C, Michel A \\
Computational Neuroscience & Baldi P, Stein C, Rinott Y, Weinshall D, Druzinsky R \\
Neural Networks & Pearlmutter B Abu-Mostafa Y, LeCun Y, \textbf{Sejnowski T}, Tang A\\
\bottomrule
\end{tabular}
\end{center}
\label{sample-table2}
\vskip -0.1in
\end{table*}

\end{document}